\documentclass{arxiv/include/bmvc_template/bmvc2k}

\usepackage{arxiv/include/bmvc_template/bmvc2k_natbib}
\usepackage{times}
\usepackage{epsfig}
\usepackage{graphicx}
\usepackage{amsmath}
\usepackage{amssymb}
\usepackage{booktabs}
\usepackage{multirow}
\usepackage{bbm}
\usepackage[normalem]{ulem}
\usepackage{rotating}

\usepackage{lipsum}
\usepackage{xr}
\usepackage{paralist}

\usepackage[font=small,labelfont=bf]{caption}

\usepackage{caption}

\definecolor{darkpastelgreen}{rgb}{0.01, 0.75, 0.24}

\title{Object-Centric Open-Vocabulary Image Retrieval with Aggregated Features}

\addauthor{Hila Levi*}{hila.levi@gm.com}{1}
\addauthor{Guy Heller*}{guy.heller@gm.com}{1}
\addauthor{Dan Levi}{dan.levi@gm.com}{1}
\addauthor{Ethan Fetaya}{ethan.fetaya@biu.ac.il}{2}

\addinstitution{
 General Motors, RND, Israel
 }
\addinstitution{
 Bar-Ilan University, Israel
}
\addinstitutionn{
\small{Equal Contribution}
}

\runninghead{H.Levi, G.Heller, D.Levi, E.Fetaya}{BMVC 2023}

\definecolor{tit_col}{rgb}{0,.1,.4}

\begin{document}

\maketitle

\begin{abstract}
   
The task of open-vocabulary object-centric image retrieval involves the retrieval of images containing a specified object of interest, delineated by an open-set text query. As working on large image datasets becomes standard, solving this task efficiently has gained significant practical importance. Applications include targeted performance analysis of retrieved images using ad-hoc queries and hard example mining during training. Recent advancements in contrastive-based open vocabulary systems have yielded remarkable breakthroughs, facilitating large-scale open vocabulary image retrieval. However, these approaches use a single global embedding per image, thereby constraining the system's ability to retrieve images containing relatively small object instances. Alternatively, incorporating local embeddings from detection pipelines faces scalability challenges, making it unsuitable for retrieval from large databases.

In this work, we present a simple yet effective approach to object-centric open-vocabulary image retrieval. Our approach aggregates dense embeddings extracted from CLIP into a compact representation, essentially combining the scalability of image retrieval pipelines with the object identification capabilities of dense detection methods.
We show the effectiveness of our scheme to the task by achieving significantly better results than global feature approaches on three datasets, increasing accuracy by up to 15 mAP points. We further integrate our scheme into a large scale retrieval framework and demonstrate our method’s advantages in terms of scalability and interpretability.

\end{abstract}


\section{Introduction}\label{sec:intro}

Retrieving images which include specific objects, according to an on-demand open-set text query, is an important task in computer vision with numerous practical applications. 
Performing such targeted searches, especially over unlabeled rare concepts, can be used, for example, to analyze the performance of an already trained system, to mine hard examples during training, or to guide the process of gathering the data for manual annotations. 
Relevant use cases vary in scale from web-scale search (e.g., Google Bard, Microsoft Bing) to 
searching in application specific datasets (e.g., e-commerce, automotive, medical applications). In both cases, scalability and efficiency 
play critical roles in adopting 
the technology.

\begin{figure}
\centering
\includegraphics[width=\linewidth]
{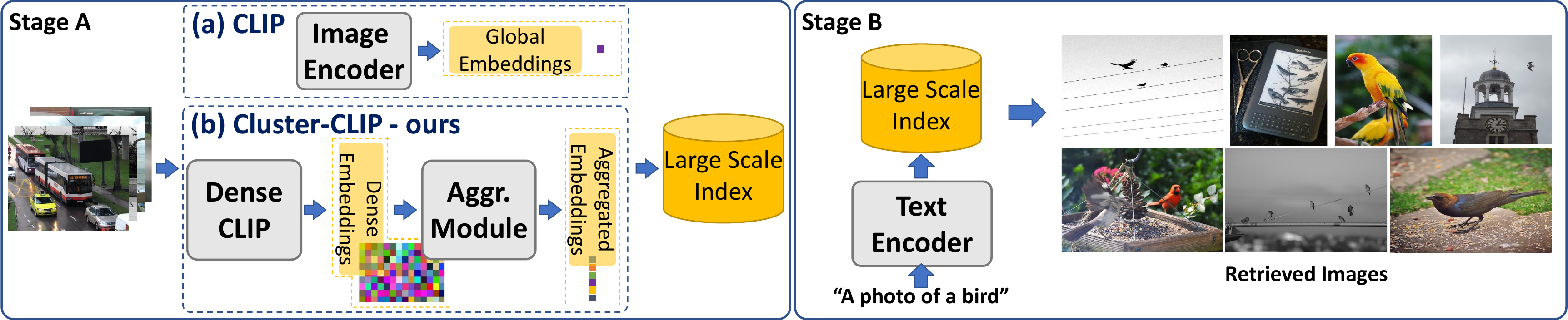}\\
\vspace{0.3cm}
\caption{\footnotesize\textbf{Overall retrieval framework}:  
Two-stage operation: Offline, we generate per-image global / aggregated embeddings via (a) CLIP or (b) Cluster-CLIP. Subsequently, an online stage enables on-demand retrieval using textual queries. Retrieval is performed by ranking text-image similarity using cosine distance in a shared embeddings space, with potential acceleration using Large Scale Index (details in Section \ref{sec:overall-framework}). This approach accommodates any dual-encoder architecture. We present top retrieved images for the query 'bird' using Cluster-CLIP features.}
\label{fig:framework}
\end{figure}

\begin{figure}[t]
\includegraphics[width=\linewidth]{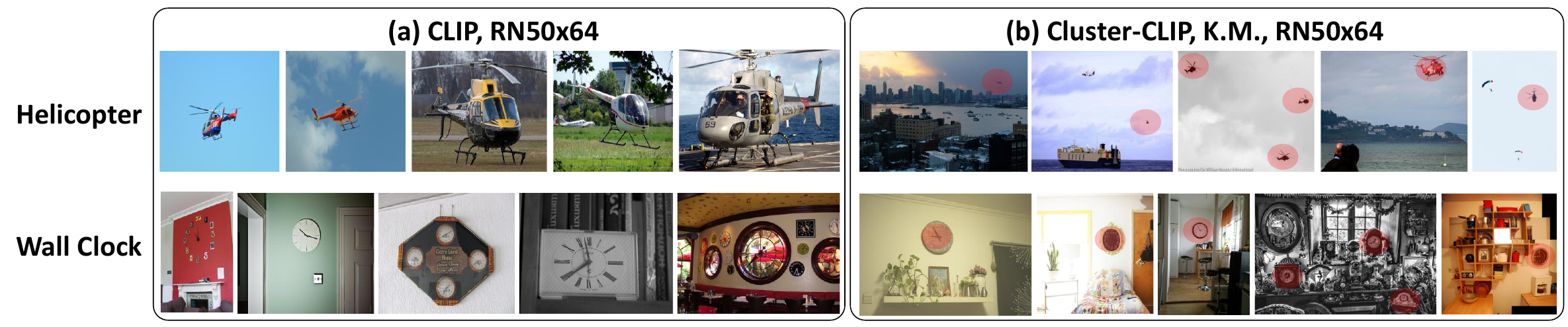}
\vspace{0.0cm}
\caption{\footnotesize\textbf{Retrieval examples} of  (a) CLIP vs. (b) Cluster-CLIP with their corresponding text queries. CLIP (single global embedding) top-scored images mainly include large objects, whereas Cluster-CLIP, which aggregates dense embeddings, also retrieves cluttered images with relatively small objects (highlighted in red circles).}
\label{fig:teaser}
\end{figure}

Despite its importance, literature lacks direct references to this task.
One possible reason might be the task complexity:
common vision-language (VL) representation of open-set objects was hard to achieve until the accelerated evolution of contrastive-based open-vocabulary models. These models (e.g., CLIP \cite{DBLP:conf/icml/RadfordKHRGASAM21}, Florence \cite{yuan2021florence}, Coca \cite{yu2022coca}), trained on web-scale image-caption data, produce a common embedding space for global image and caption representations, maximized directly via training. Retrieval is then performed by ranking the text-image similarity using cosine distance in the common embeddings space and can be scaled by using frameworks as schematically illustrated in Figure \ref{fig:framework}. 
However, empirical experiments with CLIP using open-set object queries on the similar task of object-centric retrieval produce less satisfactory results (see Figure  \ref{fig:teaser} and direct quantitative comparison in Section \ref{sec:experiments}). 
In particular, performance degrades with the increase in image complexity and the decrease in relative objects sizes.
Alternatively, the use of pure detections from SoTA open-vocabulary detection frameworks (e.g., OwlViT \cite{owlvit}) is more compatible with cluttered images but 
is considered ill-suited for retrieval tasks;  Running the detection model on each image for each query will require an enormous amount of computational resources, while precomputing and saving its internal dense embeddings will require orders of magnitude more storage.

In this work, we visit the task of object-centric open-vocabulary image retrieval and present  a simple approach to tackle it based on the complementary advantages of classification and detection open-vocabulary frameworks. The main challenge in this task is to combine the scalability of image retrieval pipelines, which can operate on huge datasets, with object-level processing of detection systems that commonly operate on a single image at a time. A second challenge is to preserve the good zero-shot accuracy obtained from web-scaled pretrained open vocabulary schemes, which holds significant importance for the task. 
We address these challenges by exploring the use of aggregated features in two steps.

As a first step towards the solution, we explore the use of local features, extracting dense embeddings from an intermediate feature-map of CLIP vision encoder and manipulating them, keeping CLIP visual-language association as is (abbreviated as Dense-CLIP). Retrieval is performed by ranking according to the maximum similarity in each image. 
In the experiments, we show that using Dense-CLIP to represent images achieves on-par results to the OwlViT baseline on all populations and significantly better results on rare objects queries, with less than half the embeddings per image. With respect to CLIP, which uses one global feature to represent each image, results are significantly increased by up to 12 mAP points. Note that while Dense-CLIP retrieval results are impressive, its use alone is not enough since it enlarges the search space by a large extent, thus impairing potential scalability.

To address scalability requirements, we explore the use of aggregated visual features by introducing Cluster-CLIP (Fig. \ref{fig:framework}). Cluster-CLIP aggregates Dense-CLIP's dense embeddings into sparse representatives with distinct local semantics (aggregation module in Fig. \ref{fig:framework}). We have examined a variety of aggregation methods that require no training to the task. By manipulating hyper-parameters of each aggregation method, we check various points on the accuracy-efficiency tradeoff (see Section \ref{sec:experiments}). Interestingly, we found out that Cluster-CLIP shows, in some cases, higher retrieval rates than Dense-CLIP (and up to 15 mAP points increase with respect to CLIP) for the small number of 10-50 representatives per image. The investigation is meaningful in that, as far as we know, it is the first work to explore and quantify the range between leveraging open vocabulary features as a global image representation and as local dense embeddings.

To summarize our contributions:

\begin{enumerate}
\vspace{-0.2cm}
\item We visit the task of object-centric open-vocabulary image retrieval and introduce Dense-CLIP, which uses CLIP's local 
features, 
keeping its original zero-shot properties.
\vspace{-0.2cm}
\item We present Cluster-CLIP which enables scalability via a compact  representation.
\vspace{-0.2cm}
\item We show the effectiveness of our approaches by achieving significantly better results compared with a global feature (CLIP) on three datasets: COCO \cite{LinMBHPRDZ14}, LVIS \cite{GuptaDG19}, and nuImages \cite{nuscenes2019}, increasing retrieval accuracy by up to 15 points.
\vspace{-0.2cm}
\item We integrate Cluster-CLIP into a retrieval framework, showcasing its scalability and presenting empirical evidence of its efficacy through plausible results. 
\end{enumerate}

\section{Related Literature}\label{sec:related}
Our work is closely related to research on instance retrieval frameworks, cross-modal retrieval, and open-vocabulary VL models. We briefly review related work in these domains.
\vspace{0.1cm}

\noindent\textbf{Instance Retrieval Frameworks.} 
Way before the deep learning era, large scale image-to-image retrieval research was dominated by mining for large-sized geographical landmarks applications, presenting methods that use local handcrafted features \cite{lowe2004distinctive, bay2008speeded} to query from  large databases of descriptors \cite{lowe2004distinctive, mikolajczyk2002affine, obdrzalek2005sub}, sometimes followed by aggregation techniques such as BoW \cite{sivic2003video}, FV \cite{jegou2011aggregating}, and VLAD \cite{jegou2010aggregating}. 
Later, at the beginning of the learning era, the use of 
learnable CNN's global descriptor \cite{arandjelovic2016netvlad, radenovic2016cnn, gordo2016deep, babenko2014neural}, local descriptors \cite{balntas2016learning, he2018local, noh2017large}, or a combination of both \cite{sarlin2019coarse, cao2020unifying, yang2021dolg} has been proposed, introducing accuracy gains with respect to the use of handcrafted features. 
Compared to these works, our application focuses on the open-vocabulary text-to-image retrieval task, yet draws inspiration from the use of local descriptors and aggregation methods that lacks in current cross modal retrieval literature.

\vspace{0.2cm}

\noindent\textbf{Cross-Modal Retrieval.} Aligning vision and language has a long-standing history of research (e.g., \cite{misrm99:mori, frome2013devise, KirosSZ14, GongWHHL14, BottomUp}). 
Among early cross-modal retrieval works, one line of research suggests learning a shared VL embedding space via a dual-path architecture \cite{VSRN, chun2021pcme, faghri2018vse++, KarpathyJL14, EngilbergeCPC18, SongS19, ThomasK20}.  Exemplars include VSRN \cite{VSRN}, which employs bottom-up attention and visual semantic reasoning, and PCME \cite{chun2021pcme}, which represents samples as probabilistic distributions. Another approach suggests using joint image-text modules \cite{SCAN, SAN, HuangWW17, LiuMLZWZ19, CAMP, MMCA, CAAN}. However, this necessitates processing all images per new text query, limiting scalability for cross-modal retrieval. A commonality across these early works is that they learn vision-language (VL) alignment using medium-sized datasets, like MS-COCO \cite{LinMBHPRDZ14} or Flicker30K \cite{Flicker30K}, resulting in decreased performance on novel datasets. 

Recently, a novel paradigm that performs Visual-Language pre-training on larger datasets has emerged \cite{ViLBERT, PixelBERT, UNITER, OSCAR, LXMERT, ALBEF}. This paradigm has catalyzed a transformative shift in the field driven by VL models pre-trained on massive web-scale corpora (CLIP \cite{DBLP:conf/icml/RadfordKHRGASAM21}, ALIGN \cite{jia2021scaling}, and others \cite{LiT, yuan2021florence, yu2022coca, SimVLM, GIT}), allowing high zero-shot performance on diverse datasets and enabling open-vocabulary retrieval. As before the VL pre-training era, some of these works employ a joint image-text module \cite{ALBEF, UNITER, OSCAR, SimVLM} that limits their scalability for cross-modal retrieval. In contrast, a more relevant line of research adopts a dual-stream approach that generates a shared embedding space for global image and caption representations, providing a straightforward solution for caption-based image retrieval tasks \cite{DBLP:conf/icml/RadfordKHRGASAM21, jia2021scaling, DeCLIP}. This line of work can be easily integrated into the retrieval scheme presented in Fig. \ref{fig:framework}. However, our study reveals that directly using global image representation for object-centric retrieval yields sub-optimal results, particularly for small-sized objects.

\vspace{0.2cm}

\noindent\textbf{Open-Vocabulary Region-Level Representation.}
This line of research employs VL pre-training for open-vocabulary object-centric tasks, such as object detection, by using region-level representation and supervision \cite{vl-survery-2307-09220}; Out of which, most relevant to us are two stream approaches \cite{cai2022x, liu2022ovis, zhou2021denseclip, GroupViT, SegCLIP, kim2023region}. Some of these works utilize a class-agnostic object detector for region proposals \cite{regionCLIP, ViLD, OpenSeg}, restricting the system's open-vocabulary capabilities. In contrast, others divide the image into a grid \cite{DBLP:conf/iclr/YaoHHLNXLLJX22, PACL, rao2022denseclip}. A representative example of the latter is OwlViT \cite{owlvit}, which is a SoTA dual-encoder open-vocabulary object detector. However, using an object detection framework directly is unfeasible for retrieval due to the need to process every image in the dataset for each new query, leading to significant computational demands. Alternatively, storing the internal embeddings of each image for the detection framework would entail a substantial increase in storage requirements, making it impractical. As our method is compatible with any dual-encoder VL open-vocabulary model, it presents a solution for leveraging progress in this line of research. For demonstration, we employ Cluster-CLIP architecture with the OwlViT backbone (see supplementary materials).

\section{Method}\label{sec:method}

The first parts of this section focus on the creation of the image embeddings 
and are organized as follows: Section \ref{subseq:preliminaries} briefly reviews the implementation details of CLIP \cite{DBLP:conf/icml/RadfordKHRGASAM21}, which produces one embedding per image. Following, Section \ref{sec:denseclip} presents Dense-CLIP\footnote{we note that this is similar yet distinct from the denseCLIP defined in \cite{zhou2021denseclip} and similarly used in \cite{rao2022denseclip} as they use a fine-tunning process and address different tasks.}, which creates dense embeddings while keeping the same embedding space as CLIP. Finally, Section \ref{sec:clustering} presents Cluster-CLIP (which sets an aggregation module on top of Dense-CLIP) and counts several clustering instantiations to create a compact representation. The last part of the section (\ref{sec:overall-framework}) presents the whole retrieval framework, as illustrated in Figure \ref{fig:framework}. 

\subsection{Preliminaries: 
CLIP}\label{subseq:preliminaries}

\begin{figure}
\centering
\includegraphics[width=0.95\linewidth]{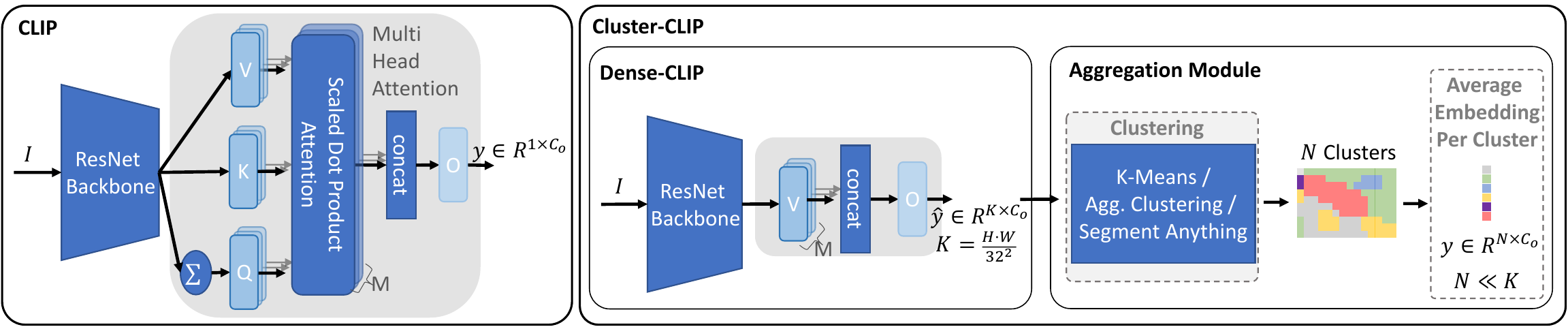}
\vspace{0.2cm}
\caption{\footnotesize\textbf{Simplified overview of image encoders}: Encoders receive image $I\in R^{H\times W \times 3}$ as input. Dense-CLIP uses only the value (V) and output (O) linear layers of CLIP's Multi-Head Attention module. Cluster-CLIP clusters Dense-CLIP output and transfers a single representative per cluster.
}
\vspace{-0.5cm}
\label{fig:vision-encoders}
\end{figure}

CLIP \cite{DBLP:conf/icml/RadfordKHRGASAM21} is a vision-language model, pretrained on massive amounts of web-scaled image-caption data via contrastive learning. It consists of two separate streams: a text-encoder, implemented as a transformer \cite{vaswani2017attention}, and a vision-encoder, implemented either by a modified ResNet backbone \cite{he2016deep} or by a ViT backbone \cite{dosovitskiy2020image}. In both cases, the last layer is a multi-head attention layer, which sums information from all the pixels in the input tensor weighted by their similarity to a query vector and projects it by an output linear layer (Fig. \ref{fig:vision-encoders}, left). The multi-head attention layer for the modified ResNet backbone is formulated as:
\begin{align}\label{eq:clip}
& y = out \left( concat \left[ y^1,  y^2, ... , y^M  \right] \right), 
& y^m = softmax \left( \frac{q^m(\bar{x}) \cdot k^m(X)^T}{\sqrt{C_q}} \right) v^m(X)
\end{align}

Here $X \in R^{K \times C_e}$ is the input tensor and $y \in R^{1 \times C_o}$ is the output embedding (one global vector of $C_o$ channels in the output representation of CLIP). 
$\bar{x}=\frac1K\sum_{i=1}^Kx_i$ represents the average of all spatial locations, $\{x_i\}_{i=1}^K$, of the input tensor $X$. $q^m: R^{C_e} \rightarrow R^{C_q}$, $k^m: R^{C_e} \rightarrow R^{C_q}$, $v^m: R^{C_e} \rightarrow R^{C_v}$ and $out: R^{MC_v} \rightarrow R^{C_o}$ are respectively the query, key, value and output linear layers, where $ m \in \{1\dots M\}$ is the index of a specific head in the multi-head architecture and $\sqrt{C_q}$ is a normalization factor. 
The described attention layer, which sums information from all spatial locations of the input tensor, is optimized to capture the "average" semantics in the image as forced to match the embedding of the corresponding caption. A reasonable hypothesis, also raised in \cite{zhou2021denseclip}, 
is that the average semantic is built upon local semantics, already captured by the spatial locations at the input to the attention layer. 

\subsection{Dense-CLIP Module}\label{sec:denseclip}

Inspired by the use of dense embeddings in detection frameworks and based on the above hypothesis, we use the following reformulation of CLIP multi-head attention layer:
\begin{align}\label{eq:denseclip}
& y_i = out \left( concat \left[ y^1_i,  y^2_i, ... , y^M_i  \right] \right),
& y_i^m = v^m(x_i)
\end{align}

Compared to the previous formulation, here the output embedding $Y \in R^{K \times C_o}$ is a tensor, and $y_i$ is the representation of its i’th spatial pixel: $\{y_i \in R^{1 \times C_o}\}_{i=1}^K$. This reformulation, implemented by removing the query and key linear layers and implementing the value and output linear layers as 1x1 convolutional layers (with the same weights), essentially creates dense patch embeddings with the same output space as CLIP (Fig. \ref{fig:vision-encoders}, middle). 
We use it as is, without fine-tuning. 
Empirical results at Section \ref{sec:experiments} reveal that Dense-CLIP 
achieves on-par retrieval accuracy as SoTA detection frameworks (i.e. OwlViT \cite{owlvit}) 
with fewer representatives. Given this finding, we next explore whether we can further reduce the number of representatives and to which extent.

\subsection{Cluster-CLIP}\label{sec:clustering}
To improve scalability, we introduce Cluster-CLIP, which produces aggregated embeddings by an additional aggregation module on top of Dense-CLIP embeddings. The aggregation module first clusters the dense features predicted by Dense-CLIP ($Y=\{y_i\}_{i=1}^K$) within $N$ clusters, denoted as $\{C_j\}_{j=1}^N$, where $C_j \subset Y$ and $N<<K$. Then, it transfers one representative embedding per cluster (the average of the embeddings within the cluster) for future retrieval use (see Figures \ref{fig:framework} and \ref{fig:vision-encoders} right). Note that the aggregation module is generally defined, and many clustering variants fit into it. We empirically examined a variety of clustering mechanisms and present here the most effective methods (while deferring the complete list and full implementation details to the Supplementary Materials):  

\vspace{0.2cm}
\noindent\textbf{K-Means (Cluster-CLIP, K.M.).}
In this method, 
we perform K-Means clustering on top of each image's dense embeddings. Once clustered, the representatives of an image are the clusters' centroids. Examples of interest that demonstrate grouping by semantic similarity can be seen in Figure  \ref{fig:clusters-example}.

\noindent\textbf{Agglomerative Clustering (Cluster-CLIP, AG.).} This method applies Agglomerative clustering (hierarchical clustering using a bottom-up approach) where the average embedding from each cluster is declared as the cluster representative. Results are presented with connectivity constraints (AG-T) and without (AG-F).

\vspace{0.2cm}
\noindent\textbf{Region Proposals (Cluster-CLIP, R.P.).} In this method, we use Segment Anything (SAM) \cite{kirillov2023segany} to segment each image. Cluster-CLIP is then provided with both the image and the masks, with the masks serving as guidance for aggregating the dense embeddings.

\subsection{Overall Framework}\label{sec:overall-framework}

A schematic illustration of our overall framework is shown in Fig. \ref{fig:framework}. Our framework consists of two separate stages. The first stage receives a dataset of images as input and uses an image encoder from a vision-language model to create embeddings through sequential processing, followed by indexing to allow a quick approximate nearest neighbour (ANN) search. In our experiments, we considered embeddings based on global embedding strategy (i.e., from CLIP), local embedding strategies (OwlViT, Ours: Dense-CLIP), and aggregations (Ours: Cluster-CLIP). The second stage receives two inputs: the large scale index created in the first stage and a textual object query wrapped by textual prompt/s. Processing includes applying the corresponding text encoder (of the same vision-language model) to create a search vector, followed by an ANN search to get a final list of ranked images. Any dual-encoder vision-language model can be integrated into this scheme. 
Notice that whereas the first stage is computationally expensive (i.e., in terms of number of FLOPS), it is executed only once. 
The above partitioning into two separate flows essentially allows us to refer to the index as given, enabling on-line interactive retrieval at the second stage. This is a significant advantage compared to object detection pipelines. The performance of the on-line system is dominated by the tradeoff between the quality of representation, the number of representatives per image, and the parameters of the ANN search. In our experiments, we follow common practice and exclude the last factor, as it is not the scope of our work, and report quantitative results based on ranking all images by cosine similarity.

\section{Experiments}\label{sec:experiments}

We evaluated our approach on the task of object-centric image retrieval 
on three publicly available datasets (COCO \cite{LinMBHPRDZ14}, LVIS \cite{GuptaDG19} and nuImages\cite{nuscenes2019}), using datasets' semantic categories as queries.
We compared the performance, in terms of retrieval accuracy vs. number of embeddings, to global and local features from existing methods. Our approach demonstrates increased performance with a small number of representatives per image, thereby allowing scalability with better retrieval rates.
 
\subsection{Datasets and Metrics}

\noindent\textbf{COCO 2017} \cite{LinMBHPRDZ14} is a very popular object detection and instance segmentation dataset of common objects in context, consisting of 120K training images and 5K validation images, fully annotated with 80 semantic categories. Categories are varied from large objects (e.g., car, elephant, tv, refrigerator) to much smaller objects (e.g., fork, book, bird, frisbee, donut).

\noindent\textbf{LVIS} \cite{GuptaDG19} is a federated dataset, which includes 20K images in its validation set, intensively used on the long-tail object detection task 
\cite{ViLD, regionCLIP, bangalath2022bridging, owlvit}. 
LVIS is annotated with 1203 semantic categories, 337 of which are considered  rare objects (less than 10 training examples). In our experiments, we separately report a retrieval metric for rare objects as an applicable approximation for the open-set retrieval task.

\noindent\textbf{nuImages} \cite{nuscenes2019} is a public
autonomous driving dataset, significantly different from the former two datasets in terms of resolution, context, RGB distribution and annotated classes. nuImages validation set includes 16.5K images, 1600x900 sized, annotated with 23 diverse semantic categories. In the experiments, we define 7 of them (those that appear in less than 0.3\% of the total annotations number) as rare categories, reporting their accuracy separately.

\noindent\textbf{Evaluation Protocol}. 
We evaluated our pipeline in two steps: a first processing step used to store the embeddings 
followed by a ranking step, which uses the datasets' categories names as queries and sorts all images in descending order of relevance per query, based on the maximal similarity over all patches. For each query, 
images are declared as \textit{true positive} if they include an object of that category. 
We report mean average precision ($mAP$), as widely reported in retrieval tasks \cite{noh2017large, cao2020unifying, yang2021dolg, ng2020solar, radenovic2018fine}, and $mAP@50$ (defined in \cite{kaggle2022gupta}), previously used in  \cite{liu2022ovis}, which considers top-k images only, as our main criteria for comparison.    

\noindent\textbf{Implementation Details}. We used the CLIP backbones from the CLIP \cite{DBLP:conf/icml/RadfordKHRGASAM21} library. 
Clustering (K-Means and hierarchical clustering) was performed via the sklearn library \cite{scikit-learn}. Region proposals were calculated by Segment Anything (SAM) \cite{kirillov2023segany} library and tuned with different number point-prompts (64, 256 and 1024 points) which created,  
approximately 25, 50 and 100 representatives per image. 
All Dense-CLIP and clustering experiments were conducted on 1 Nvidia GPU machine. Images were resized to a square aspect ratio following hyperparameter search, and positional embeddings were interpolated to match the resolution of the image. For a fair comparison, we ensemble over the embeddings space of the 7 best CLIP prompts \cite{owlvit} in all VL pre-trained modules. 
Full architectures descriptions, design choices, and hyperparameters are specified in the supplementary.

\noindent\textbf{Baselines}. 
We compare our work with existing global embeddings from CLIP and local embeddings from OwlViT \cite{owlvit}. As described in Section \ref{sec:related}, OwlViT is an 
open-vocabulary dual-branch VL detection model that achieves SoTA results, making it a strong baseline for our task. It is pre-trained on a web-scale dataset and then fine-tuned for open-vocabulary object detection, which can lead to a forgetting effect. For image representation, we directly used local embeddings from OwlViT's ViT \cite{dosovitskiy2020image} vision-encoder output, excluding bounding boxes prediction. We additionally compare to dual-stream cross-modal retrieval methods preceding CLIP that are trained on distinct medium-size datasets using caption annotations. Specifically, we compare to PCME and VSRN trained on COCO using the split defined in \cite{KarpathyJL14} and to VSRN trained on Flicker30K \cite{Flicker30K} (VSRN is referred to as either `VSRN, COCO-Caption' or `VSRN, Flicker30K', depending on the dataset).

\begin{table}[t]
\begin{center}
\resizebox{\textwidth}{!}{%
\begin{tabular}{|lcc|lll|ll|ll|}
\hline
& & & \multicolumn{3}{|c|}{COCO} & \multicolumn{2}{|c|}{LVIS} & \multicolumn{2}{|c|}{LVIS-rare}\\
Backbone & Res. & $\#$rep. & mAP@50 & mAP & mAP@50$_{s-m}$ & mAP@50 & mAP & mAP@50 & mAP\\
\hline\hline
VSRN, Flicker30K & 600 & 1 & 44.28 & 37.23 & 21.49 & 31.56 & 37.35 & 10.52 & 11.53 \\
VSRN, COCO-Caption & 600 & 1 & {70.29} & {52.33} & {30.99} & 42.09 & 45.60 & 20.46 & 21.36 \\
PCME & 224 & 7 & {69.69} & {57.98} & {29.98} & 47.38 & 51.06 & 27.85 & 28.58 \\
\hline\hline
CLIP, RN50 & 224 & 1 & 56.70 & 50.80 & 21.91 & 52.69 & 55.94 & 39.84 & 40.74 \\
CLIP, RN50x4 & 288 & 1 &64.58 & 56.39 & 29.39 & 57.85 & 60.35 & 43.37 & 44.28 \\
CLIP, RN50x64 & 448 & 1 & 70.62 & 61.03 & 36.60 & 62.60 & 64.71 & 53.14 & 53.92\\
\hline\hline
OwlViT, ViT-B/32 & 768 & 576 & 77.31 & \textcolor{blue}{70.37} & \textcolor{blue}{52.22} & 66.09 & 67.86 & 42.60 & 42.95 \\
OwlViT, ViT-B/16 & 768 & 2304 & 74.96 & 65.91 & 47.02 & 61.28 & 63.39 & 34.75 & 35.82 \\
OwlViT, ViT-L/14 & 840 & 3600 & 76.61 & \textcolor{red}{71.06}
 & \textcolor{red}{58.95} &66.15 & 67.86 & 40.62 & 41.10 \\
\hline\hline
Dense-CLIP, RN50 & 512 & 256 & 58.83 {\scriptsize(+2.13)} & 52.35 {\scriptsize(+1.55)} & 32.58 {\scriptsize(+10.67)}  &55.41 {\scriptsize(+2.72)} & 57.46 {\scriptsize(+1.52)} & 38.80 {\scriptsize(-1.04)} & 39.86 {\scriptsize(-0.88)} \\
Dense-CLIP, RN50x4 & 512 & 256 & 69.61 {\scriptsize(+5.03)} & 62.10 {\scriptsize(+5.71)} & 41.18 {\scriptsize(+11.79)} & 63.88 {\scriptsize(+6.03)} & 65.88 {\scriptsize(+5.53)} & 55.32 {\scriptsize(+11.95)} & 56.40 {\scriptsize(+12.12)} \\
Dense-CLIP, RN50x64 & 448 & 196 & 77.78 {\scriptsize(+7.16)} & 69.65 {\scriptsize(+8.62)} & 51.47 
 {\scriptsize(+14.87)} &70.86 {\scriptsize(+8.26)} & 71.80 {\scriptsize(+7.09)} & 57.97 {\scriptsize(+4.83)} & 58.72 {\scriptsize(+4.80)} \\
\hline\hline
CLIP + Dense-CLIP, RN50 & 512 & 257 & 67.27 {\scriptsize(+10.57)}  & 59.33 {\scriptsize(+8.53)} & 34.99 {\scriptsize(+13.08)} & 61.32 {\scriptsize(+8.63)} & 63.02 {\scriptsize(+7.08)} & 46.96 {\scriptsize(+7.12)} & 47.82 {\scriptsize(+7.08)} \\
CLIP + Dense-CLIP, RN50x4 & 512 & 257 & 66.21 {\scriptsize(+1.63)} & 58.08 {\scriptsize(+1.69)} & 29.80 {\scriptsize(+0.41)} & 61.82 {\scriptsize(+3.97)}  & 64.12 {\scriptsize(+3.77)} & 49.49 {\scriptsize(+6.12) } & 50.35 {\scriptsize(+6.07)} \\ 
CLIP + Dense-CLIP, RN50x64 & 448 & 197 & 77.48 {\scriptsize(+6.86)} & 69.45 {\scriptsize(+8.42)} & 48.73 {\scriptsize(+12.13)} & \textcolor{red}{72.24} {\scriptsize(+9.64)}  & \textcolor{red}{73.36} {\scriptsize(+8.65)} & \textcolor{red}{61.68} {\scriptsize(+8.54)} & \textcolor{red}{62.33} {\scriptsize(+8.41)} \\
\hline\hline
Cluster-CLIP, K.M., RN50x64 & 448 & 10 & \textcolor{blue}{78.67} {\scriptsize(+8.05)} & 63.66 {\scriptsize(+2.63)} & 46.69 {\scriptsize(+10.09)} &  62.72 {\scriptsize(+0.12)}  & 64.11 {\scriptsize(-0.6)} & 51.66 {\scriptsize(-1.48)} & 52.45 {\scriptsize(-1.47)} \\
Cluster-CLIP, AG-T., RN50x64 & 448 & 10 & 76.72 {\scriptsize(+6.10)} & 64.00 \scriptsize(+2.97) & 45.03 \scriptsize(+8.43) & 63.70 \scriptsize(+1.1) & 65.16 \scriptsize(+0.45) & 49.53 \scriptsize(-3.61) & 50.38 \scriptsize(-3.54) \\
Cluster-CLIP, AG-F., RN50x64 & 448 & 50 & 78.35 \scriptsize(+7.73) & 69.84 \scriptsize(+8.81) & 51.95 \scriptsize(+15.35) & {71.63} \scriptsize(+9.03) & {72.60} \scriptsize(+7.89) & {58.29} \scriptsize(+5.15) & {59.07} \scriptsize(+5.15)\\
Cluster-CLIP, R.P., RN50x64 & 448 & 91 & \textcolor{red}{79.24} {\scriptsize(+8.62)} & 69.43 {\scriptsize(+8.4)} & 50.51 {\scriptsize(+13.91)} & 70.74 {\scriptsize(+8.14)}  & 71.92 {\scriptsize(+7.21)} & 58.28 {\scriptsize(+5.14)} & 58.88 {\scriptsize(+4.96)} \\
\hline\hline
CLIP + Cluster-CLIP, K.M, RN50x64 & 448 & 11 & 75.51 \scriptsize(+4.89) & 64.60 \scriptsize(+3.57) & 41.61 \scriptsize(+5.01) & 66.39 \scriptsize(+3.79) & 67.96 \scriptsize(+3.25) & 57.02 \scriptsize(+3.88) & 57.73 \scriptsize(+3.81) \\
CLIP + Cluster-CLIP, AG-T, RN50x64 & 448 & 11 & 75.42 \scriptsize(+4.8) & 65.19 \scriptsize(+4.16) & 43.22 \scriptsize(+6.62) & 67.16 \scriptsize(+4.56) & 68.64 \scriptsize(+3.93) & 57.15 \scriptsize(+4.01) & 57.85 \scriptsize(+3.93) \\
CLIP + Cluster-CLIP, AG-F, RN50x64 & 448 & 51 & 77.06 \scriptsize(+6.44) & 69.03 \scriptsize(+8.00) & 48.08 \scriptsize(+11.48) & \textcolor{blue}{71.79} \scriptsize(+9.19) & \textcolor{blue}{73.02} \scriptsize(+8.31) & {60.92} \scriptsize(+7.78) & \textcolor{blue}{61.61} \scriptsize(+7.69) \\
CLIP + Cluster-CLIP, R.P., RN50x64 & 448 & 92 & 77.75 \scriptsize(+7.13) & 68.59 \scriptsize(+7.56) & 46.75 \scriptsize(+10.15) & 71.26 \scriptsize(+8.66) & 72.59 \scriptsize(+7.88) & \textcolor{blue}{60.98} \scriptsize(+7.84) & 61.21 \scriptsize(+7.29) \\
\hline
\end{tabular}}
\end{center}
\caption{\footnotesize Evaluation results on COCO2017 and LVIS val sets. \textcolor{red}{First} and \textcolor{blue}{second} best scores are marked in red and blue. Using Dense-CLIP improves retrieval accuracy but increases the number of features. Using Cluster-CLIP compensates both, enabling scaling.}
\label{tab:dense_cocolvis}
\vspace{-0.2cm}
\end{table}

\subsection{Results}

\begin{figure}[h]
	\begin{minipage}[t]{\textwidth}
		\begin{minipage}[]{0.73\textwidth}
\begin{center}
\resizebox{.95\linewidth}{!}{%
\begin{tabular}{|lcc|ll|ll|}
\hline
& & & \multicolumn{2}{|c|}{nuImages} & \multicolumn{2}{|c|}{nuImages - rare} \\
Backbone & res. & $\#$rep. & mAP@50 & mAP & mAP@50 & mAP \\
\hline\hline
PCME & 224 & 7 & 25.67 & 15.08 & 0.2 & 0.75 \\
\hline\hline
CLIP, RN50 & 224 & 1 & 27.20 & 17.61 & 0.84 & 1.58 \\
CLIP, RN50x4 & 288 & 1 & 28.62 & 19.07 & 2.70 & 3.39\\
CLIP, RN50x64 & 448 & 1 & 31.93 & 20.77 & 3.16 & 4.28 \\
\hline\hline
OWL-ViT, ViT-B/32 & 768 & 576 & 36.93 & 27.19 & 1.88 & 2.48 \\
OWL-ViT, ViT-B/16 & 768 & 2304 & 33.45 & 26.65 & 2.25 & 2.58 \\
OWL-ViT, ViT-L/14 & 840 & 3600 & 30.30 & 25.10 & 2.12 & 3.43 \\
\hline\hline
Dense-CLIP, RN50 & 768 & 576 & 31.64 {\scriptsize(+4.44)} & 24.93 {\scriptsize(+7.32)} & 3.33 {\scriptsize(+2.49)} & 4.95 {\scriptsize(+3.37)} \\
Dense-CLIP, RN50x4 & 768 & 576 & 32.04 {\scriptsize(+3.42)} & 26.77 {\scriptsize(+7.70)} & 2.80 {\scriptsize(+0.10)} & 4.88 {\scriptsize(+1.49)} \\
Dense-CLIP, RN50x64 & 768 & 576 & 34.08 {\scriptsize(+2.15)} & {30.74} {\scriptsize(+9.97)} & \textcolor{blue}{10.11} {\scriptsize(+6.95)} & \textcolor{blue}{10.88} {\scriptsize(+6.60)} \\
\hline\hline
CLIP + Dense-CLIP, RN50 & 768 & 577 & 32.60 {\scriptsize(+5.40)} & 25.68 {\scriptsize(+8.07)} & 3.41 {\scriptsize(+2.57)} & 5.13 {\scriptsize(+3.55)}\\
CLIP + Dense-CLIP, RN50x4 & 768 & 577 & 33.79 {\scriptsize(+5.17)} & 25.94 {\scriptsize(+6.87)} & 6.32 {\scriptsize(+3.62)} & 7.70 {\scriptsize(+4.31)} \\
CLIP + Dense-CLIP, RN50x64 & 768 & 577 & 37.95 {\scriptsize(+6.02)} & \textcolor{red}{31.26} {\scriptsize(+10.49)} & \textcolor{red}{10.23} {\scriptsize(+7.07)} & \textcolor{red}{11.24} {\scriptsize(+6.96)} \\
\hline\hline
Cluster-CLIP, K.M., RN50x64 & 768 & 10 & 37.87 {\scriptsize(+5.94)} & 25.02 {\scriptsize(+4.25)} & 6.07 {\scriptsize(+2.91)} & 6.93 {\scriptsize(+2.65)} \\ 
Cluster-CLIP, AG-T., RN50x64 & 768 & 10 & \textcolor{red}{40.55} \scriptsize(+8.62) & 27.28 \scriptsize(+6.51) & 7.35 \scriptsize(+4.19) & 8.51 \scriptsize(+4.23) \\

Cluster-CLIP, AG-F., RN50x64 & 768 & 50 & 37.00 \scriptsize(+5.07) & 30.61 \scriptsize(+9.84) & 9.32 \scriptsize(+6.16) & 10.40 \scriptsize(+6.12) \\
Cluster-CLIP, R.P., RN50x64 & 768 & 97 & {40.05} {\scriptsize(+8.12)} & 30.05 {\scriptsize(+9.28)} & 9.10 {\scriptsize(+5.94)} & 10.16 {\scriptsize(+5.88)} \\
\hline\hline
CLIP + Cluster-CLIP, K.M., RN50x64 & 768 & 11 & 39.78 \scriptsize(+7.85) & 26.95 \scriptsize(+6.18) & 8.13 \scriptsize(+4.97) & 9.15 \scriptsize(+4.87) \\
CLIP + Cluster-CLIP, AG-T, RN50x64 & 768 & 11 & 40.24 \scriptsize(+8.31) & 28.06 \scriptsize(+7.29) & 8.82 \scriptsize(+5.66) & 9.80 \scriptsize(+5.52) \\
CLIP + Cluster-CLIP, AG-F, RN50x64 & 768 & 50 & 38.55 \scriptsize(+6.62) & \textcolor{blue}{30.75} \scriptsize(+9.98) & 9.10 \scriptsize(+5.94) & 10.37 \scriptsize(+6.09) \\
CLIP + Cluster-CLIP, R.P., RN50x64 & 768 & 98 & \textcolor{blue}{40.54} \scriptsize(+8.61) & 29.84 \scriptsize(+9.07) & 8.84 \scriptsize(+5.68) & 10.15 \scriptsize(+5.87) \\ 
\hline 
\end{tabular}}
\end{center}

		\end{minipage}
		\hfill
		\begin{minipage}[]{0.25\textwidth}
			\centering
			\includegraphics[width=0.78\textwidth]{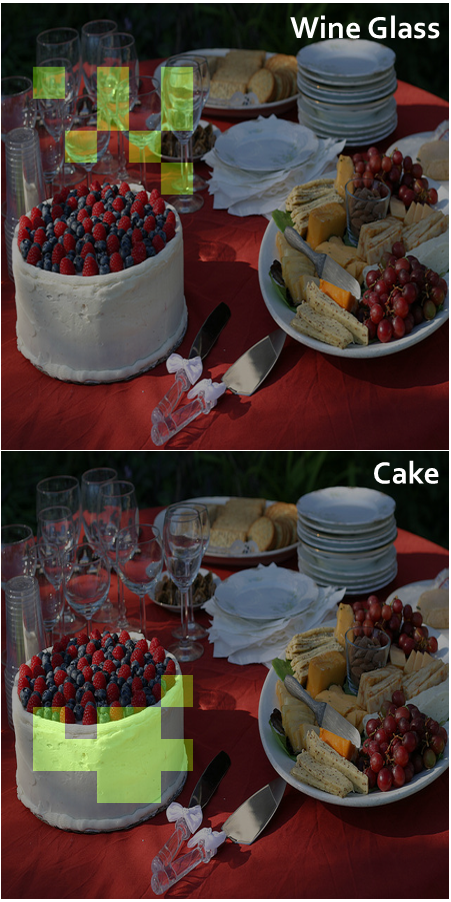}
		\end{minipage}\\
		\begin{minipage}[t]{0.73\linewidth}
			\centering
			\captionsetup{width=\textwidth}
			\captionof{table}{\footnotesize Evaluation results on nuImages val set. \textcolor{red}{First} and \textcolor{blue}{second} best scores are \\ marked in red and blue.}
			\label{tab:dense_nuimages}
		\end{minipage}
		\centering
		\begin{minipage}[t]{0.25\textwidth}
			\centering
			\captionsetup{width=\textwidth}
			\captionof{figure}{\footnotesize Top matching clusters.}
			\label{fig:clusters-example}
		\end{minipage}
	\end{minipage}
 \vspace{-0.3cm}
\end{figure}

\noindent\textbf{Dense-CLIP Results.} Tables \ref{tab:dense_cocolvis} and \ref{tab:dense_nuimages} compare the retrieval results of Dense-CLIP to the baselines. The '$\#$rep.’ column shows the average number of embeddings for each image. Dense-CLIP achieves on-par results compared to OwlViT with fewer embeddings per image and leads to a significant and consistent improvement of up to 12 points in the retrieval rates over CLIP. Furthermore, Dense-CLIP surpasses VSRN and PCME on COCO and, to a greater extent, on LVIS, even when the latter two were fine-tuned on these datasets' images. 
Small object retrieval ($mAP@50_{s-m}$ in Table \ref{tab:dense_cocolvis}) benefits from the use of local features, where Dense-CLIP achieves competitive results with respect to OwlViT. Rare
object retrieval (right side of the tables) proved to be more difficult (lower accuracy). Interestingly, results show higher retrieval rates for CLIP over OwlViT, maybe because of forgetting effects due to finetuning. Dense-CLIP shows a significant improvement over all baselines, exploiting CLIP open vocabulary representations in a dense manner.

\begin{figure}
\includegraphics[width=\linewidth]{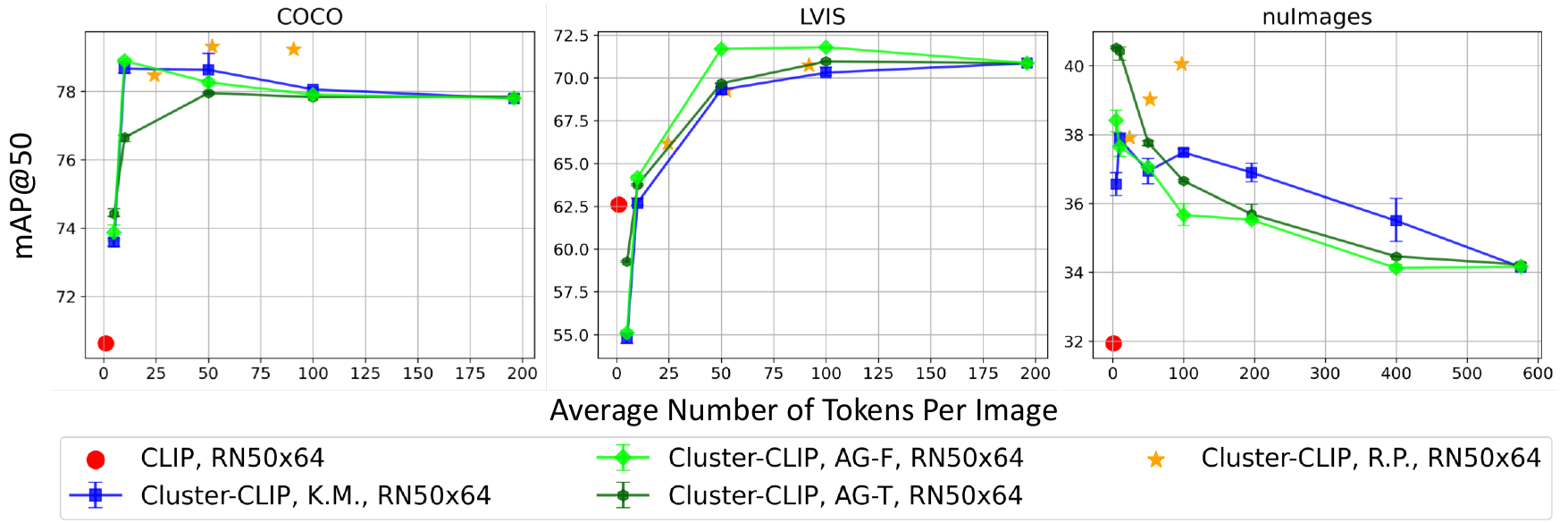}\\
\caption{\textbf{Cluster-CLIP accuracy-efficiency scatter plots:} retrieval accuracy (mAP@50) vs. average numbers of embeddings per image for COCO, LVIS and nuImages datasets. Top left is better. Cluster-CLIP achieves high accuracy for the small number of 10-50 embeddings per image.} 
\vspace{-0.4cm}
\label{fig:clustering_res}
\end{figure}

\vspace{\baselineskip}
\noindent\textbf{Cluster-CLIP Results.}
We empirically studied the use of aggregated features using several clustering mechanisms, described in section \ref{sec:clustering}. Fig. \ref{fig:clustering_res} displays Cluster-CLIP's results for different clustering methods and numbers of clusters, while promising working points are presented in the sixth parts of Tab. 
\ref{tab:dense_cocolvis} and \ref{tab:dense_nuimages} (see results for RN50/x4 backbones in the supplementary). To reduce variance, all results are averaged across three experiments. 

Notably, K.M. and AG-F (blue and lime) outperform Dense-CLIP when employing a modest count of 10 representatives per image on COCO. With a larger representatives budget of 50, aggregating the dense features according to R.P. (star markers) is made an alternative clustering mechanism. On LVIS, which has almost twice the number of instances per image compared to COCO, AG-F with 50 representatives achieves the highest retrieval rates. Interestingly, while AG-F and K.M. share certain clustering characteristics, it appears that AG-F bottom-up clustering is preferred when fine-grained understanding is required (as in LVIS) and on small objects. On nuImages, AG-T and R.P., which incorporate localization considerations into the clustering process, outperform other mechanisms with merely 5-50 representatives, implying localization is a strong cue for clustering in high-resolution images. 
As mentioned, Cluster-CLIP outperforms Dense-CLIP in several cases. This can be explained by noticing that applying clustering to the output of Dense-CLIP introduces two conflicting phenomena. On the one hand, clustering reduces the number of representatives per image, thereby decreasing the presence of distractors. On the other hand, averaging across multiple patches may lead to the averaging out of small-sized objects or fine-grained details, potentially hindering performance. 

We consider the above results as key contributions, highlighting the potential of aggregated features to offer both efficiency and performance.

\vspace{\baselineskip}
\noindent\textbf{Mixed Architectures.}
We found that in many cases, Dense-CLIP and Cluster-CLIP embeddings are in fact complementary to the global embedding produced by CLIP, and those can be effectively combined to boost their performance (referred to as CLIP + Dense/Cluster-CLIP). Results are reported in the fifth and seventh part of Tables \ref{tab:dense_cocolvis} and \ref{tab:dense_nuimages}. Notably, incorporating CLIP embeddings enhances Dense-CLIP results by up to 3.8 mAP@50 points. Moreover, integrating CLIP embeddings into Cluster-CLIP boosts results by up to 7.6 mAP@50 points while reinforcing its efficacy across tasks.

\subsection{Qualitative Results with Overall Framework}\label{sec:exp_qual}

\begin{figure*}[b]
\includegraphics[width=\linewidth]{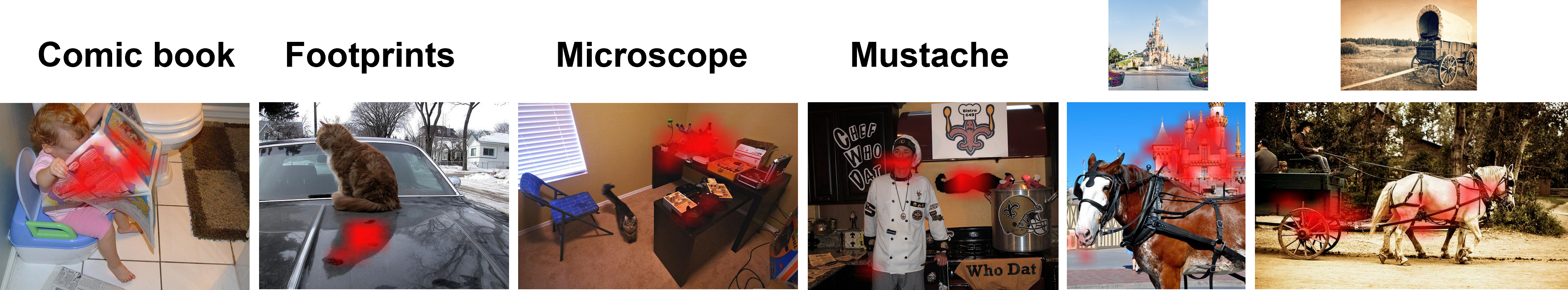}\\
\caption{\textbf{Qualitative Examples}: top-1 retrieved images with Dense-CLIP heat-maps for rare textual and visual queries. Index created from 120K images with K.M. clustered embeddings.}
\label{fig:qualitative}
\vspace{-0.3cm}
\end{figure*}

We build a simple image retrieval framework to study the behavior of Cluster-CLIP aggregated embeddings (Figure \ref{fig:framework}). For demonstration purposes, we use an index consisting of 120K COCO training set images. Specifically, we used Cluster-CLIP with RN50x64 backbone and K-Means with 10 centroids, then utilized FAISS \cite{johnson2019billion} to map the embeddings into a large-scale index. Interactive querying is enabled using text or image queries (created by CLIP text or image encoder). 
Qualitative examples of interest 
are demonstrated in Figures \ref{fig:framework}, \ref{fig:teaser} and  \ref{fig:qualitative}. Please refer to the Supplementary Materials for additional qualitative examples. 

\vspace{-0.3cm}
\section{Conclusions}\label{sec:conclusions}
We examine the possible use of local features for the object-centric 
image retrieval task and introduce Dense-CLIP, which extracts dense embedding, manipulated such that CLIP vision-language association is kept. 
We compare our method with the use of global features extracted from CLIP with the same backbone and report a significant increase in retrieval rates of up to 12 mAP points. Compared to SoTA detection pipelines, our results are competitive, with a significant increase in retrieval rates for rare categories.

Practically, utilizing local features within a retrieval framework significantly enlarges the search space, impairing scalability. To address this, we introduce Cluster-CLIP, which innovatively represents images using features aggregated from local features. 
Our approach achieves improved retrieval rates with fewer representatives, practically enabling scaling. From a broader perspective, the potential use of compact representation to efficiently carry useful image information is interesting by itself and might contribute to future work on a wide range of applications (e.g., detection, segmentation, image generation).


\pagebreak
\vskip\baselineskip
\vskip\baselineskip
\begin{raggedright}
    \textbf{\LARGE \bfseries\sffamily\textcolor{tit_col}{Supplementary Materials}}
\end{raggedright}
\vskip\baselineskip

\setcounter{equation}{0}
\setcounter{figure}{0}
\setcounter{table}{0}
\setcounter{page}{1}
\setcounter{section}{0}
\renewcommand{\thesection}{S\the\value{section}}
\renewcommand{\theHsection}{S\the\value{section}}
\renewcommand{\thefigure}{S\the\value{figure}}
\renewcommand{\theequation}{S\the\value{equation}}

\section{Clustering Methods}\label{app:clustering_methods}
In our work, we introduce Cluster-CLIP, a method that represents images using a compact representation by adding an aggregation module on top of Dense-CLIP dense emebeddings (as detailed in Section \ref{sec:clustering} of the main article). The aggregation module first clusters the dense embeddings and then transfers a single representative per cluster. We empirically evaluated various clustering methods within the aggregation module, which are presented in this section, with their results reported in Section \ref{app:experiments}.

\vspace{0.2cm}
\noindent\textbf{K-Means (K.M.).}
In this method, we perform K-Means clustering on top of each image's dense embeddings. Once clustered, the representatives of an image are the clusters' centroids. We hypothesize that in such a way, each group of semantically similar objects will be represented by their common semantics. An example of such behavior can be seen in Figure \ref{fig:clusters-example} in the main article, where several wine glasses are represented by a single cluster, which also scores highest among the different clusters when compared to the embeddings of the phrase "Wine Glass".

\vspace{0.2cm}
\noindent\textbf{Agglomerative Clustering (AG).} 
This method applies Agglomerative clustering, which performs hierarchical clustering using a bottom-up approach: each observation starts in its own cluster, and clusters are successively merged together via a linkage criterion. Different linkage criteria were tested, with and without connectivity constraints; Specifically, Ward linkage, which minimizes the sum of squared differences within all clusters, and Average linkage, which minimizes the average of the distances between all observations of pairs of clusters. Average linkage was tested both with Euclidean and Cosine metrics. We found that using Ward linkage works better, and so we present its results in Section \ref{app:experiments} with connectivity constraints, marked AG-T, and without, marked AG-F. 

\vspace{0.2cm}
\noindent\textbf{Region Proposals (R.P.).}
In this method, we use Segment Anything (SAM) \cite{kirillov2023segany} to segment each image. Then, Cluster-CLIP is provided with both the image and the masks, with the masks serving as guidance for clustering the dense embeddings. Formally, given an image of dimension $H \times W$, and the matching Dense-CLIP dense embeddings of dimensions $\frac{H}{32} \times \frac{W}{32} \times C_o$, where $C_o$ is the number of channels at CLIP's output. For each binary mask $m \in  R^{H \times W}$ predicted by SAM, we first use max pooling to down-sample the mask to the dense embeddings resolution. Then, we aggregate dense embeddings which coincide with the downsampled mask. Once clusters are formed, each cluster is represented with the mean of its embeddings. To adjust the final number of masks per image, we conducted experiments with various quantities of candidate point-prompts to SAM. Specifically, we ran with 64, 256, and 1024 candidates, which resulted in different numbers of masks per image.

\vspace{0.2cm}
\noindent\textbf{Soft Aggregation via Attention (AT).} The clustering algorithms in our work, and K-Means specifically, aggregate each cluster's embeddings by taking the mean over them. Therefore, each cluster representative includes information aggregated only in its cluster (Hard Aggregation). In this method, we suggest weighted aggregation of non-local information, i.e., from all of the image embeddings. This idea is implemented by adapting the attention mechanism described in Section \ref{subseq:preliminaries} in two steps: \begin{inparaenum}[1)]
\item Clusters are computed on the inputs to the attention layer.
\item The means of the clusters' embeddings (centroids for K-Means) are used as queries in the attention mechanism. 
\end{inparaenum}

Using the notations from eq. \ref{eq:clip} in the main article, this can be formulated as:
\begin{align}\label{eq:attn-denseclip}
& y_j = out \left( concat \left[ y_j^1,  y_j^2, ... , y_j^M  \right] \right) \nonumber \\
& y^m_j = softmax \left( \frac{q^m(c_j) \cdot k^m(X)^T}{\sqrt{C_q}} \right) v^m(X)
\end{align}

Here $c_j$ and $y_j$ are the mean and soft aggregated representation of the j'th cluster, respectively: $\{c_j\in R^{1 \times C_e}\}_{i=1}^N$, $\{y_j\in R^{1 \times C_o}\}_{i=1}^N$, where $N$ is the number of clusters. This reformulation inherits information from the clustering mechanism (here K-Means) and uses CLIP pretrained query, key, and value weights to essentially create aggregated embeddings with the same output space as CLIP, keeping its zero-shot performance.

\vspace{0.2cm}
\noindent\textbf{Adaptive K-Means (A-K.M.).} As different images can contain different numbers of categories, applying K-Means with an adaptive number of clusters per image as a function of the image properties might also be beneficial. A-K.M. uses the Bayesian information criterion (BIC) \cite{schwarz1978estimating}, which is a popular criterion for model selection, in an attempt to choose the best number of clusters per image. 

The BIC score of a probabilistic model $Q$ is defined as 
\begin{equation}
    BIC(Q) = \kappa\ln(n) - 2\ln(\hat{L})
\end{equation}
Here, $\kappa$ is the number of estimated parameters in $Q$, $n$ is the number of samples observed, and $\hat{L}$ is the model's maximized likelihood function for the observed samples. A lower BIC value is commonly considered better, as it balances the model's complexity (in terms of the number of parameters) and the model fit. To that end, the term $\kappa\ln(n)$ functions as a penalty against utilizing models with a larger number of parameters in order to inflate the likelihood of the model. 

To apply a BIC score for K-Means, the method models K-Means with $k$ clusters as a Gaussian Mixture Model (GMM) 
with $k$ components and spherical covariance. Each GMM component represents a cluster by setting the component's mean to the cluster's centroid and estimating the covariance by the cluster's embeddings.

Using the above definition of BIC score for K-Means, the following algorithm is used to select the best number clusters. Let ${k_1, k_2, ... k_n}$ be a collection of choices for the number of clusters selected apriori, such that $k_i < k_{i+1}$, and mark by $BIC_{k_i}$ the BIC score computed over clusters produced by K-Means with $k_i$ clusters. If $\exists k_i: BIC_{k_i} < BIC_{k_{i+1}}$, then $k_i$ is selected as the number of clusters for the image; otherwise $k_n$ is selected.

\vspace{0.2cm}
\noindent\textbf{Anchors (AN).} In this method, the dense representations are clustered according to a spatial division. The resized image is divided into equal-sized squares in multiple resolutions, and the matching embeddings at each resolution are clustered together. The embeddings in each cluster are averaged to create a single representative per cluster.

\begin{figure}[b!]
\includegraphics[width=\linewidth]{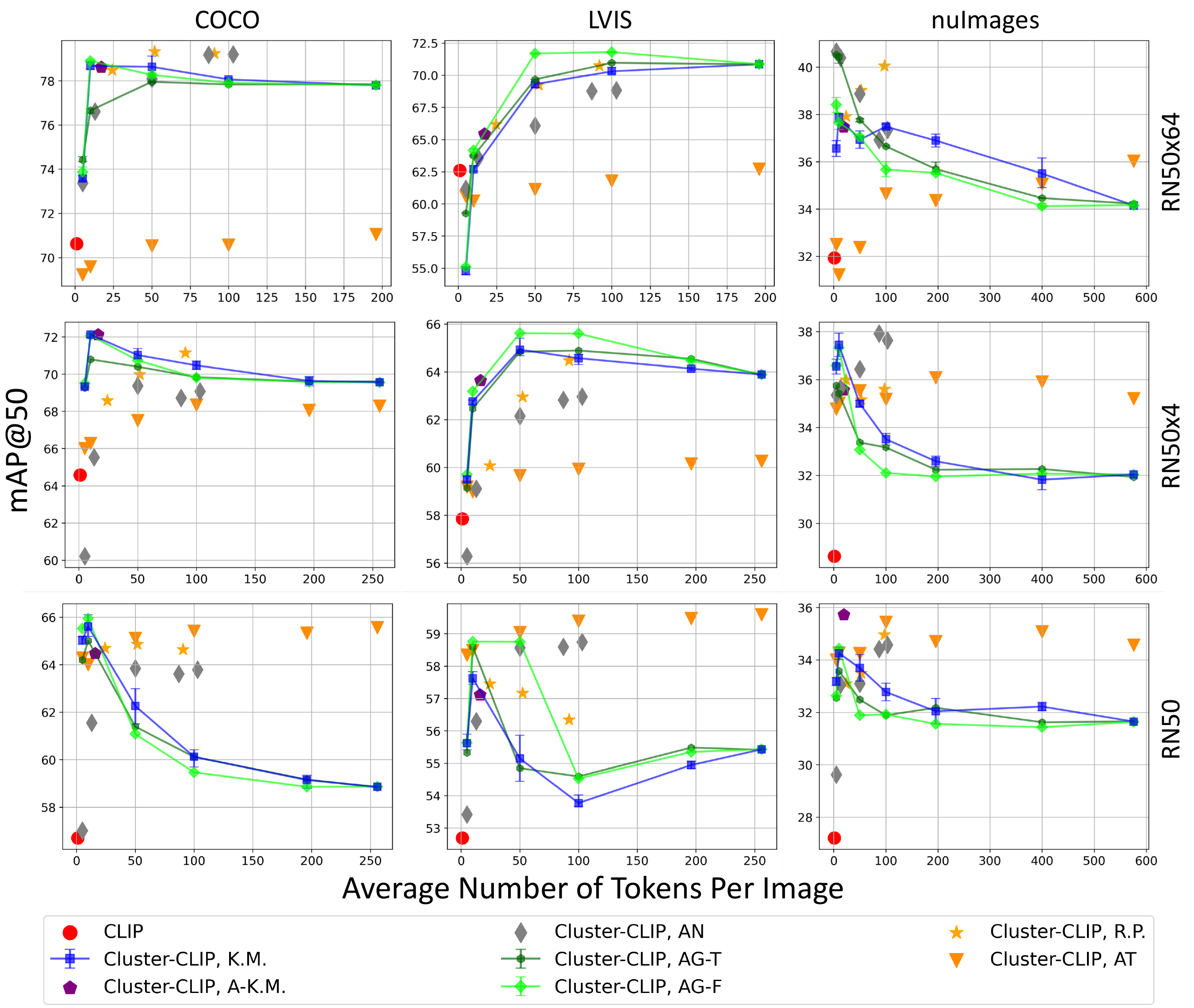}\\
\caption{\textbf{Cluster-CLIP accuracy-efficiency scatter plots for different clustering methods:} retrieval accuracy (mAP@50) vs. average numbers of embeddings per image for COCO, LVIS, and nuImages datasets. Top left is better.} 
\vspace{-0.2cm}
\label{app:fig:clustering_res_sup}
\end{figure}

\section{Cluster-CLIP Results}\label{app:experiments}
This section extends Cluster-CLIP results from Section \ref{sec:experiments} of the main article by evaluating the additional clustering methods described in Section \ref{app:clustering_methods}, and extending the results for other ResNet backbones.
Figure \ref{app:fig:clustering_res_sup} presents the results in terms of retrieval accuracy (mAP@50) vs. average number of embeddings per image using RN50x64, RN50x4, and RN50 backbones (first, second, and third rows) on COCO \cite{LinMBHPRDZ14}, LVIS \cite{GuptaDG19}, and nuImages \cite{nuscenes2019} datasets (left, middle and right columns). K-Means and Agglomerative Clustering are presented by blue and green solid lines, while other clustering methods are depicted by scatter plots. CLIP is denoted by red circles.

From Figure \ref{app:fig:clustering_res_sup}, we can see the effectiveness Cluster-CLIP top-performing clustering methods mentioned in Section \ref{sec:clustering} of the main article. Specifically, K-Means (K.M.), Agglomerative Clustering (AG-T/F), and Region Proposals (R.P) outperform CLIP when using the same backbone architecture across all datasets with merely 5-50 representatives per image, showcasing Cluster-CLIP effectiveness across backbones.
When considering the Adaptive K-Means method (purple pentagon), we see that adaptively selecting the number of clusters often scores close to the interpolated score of K-Means with no significant gain.  
For the Anchors method (gray diamond), it is generally advantageous to partition the embedding space into a greater number of scales or utilize finer divisions, thereby increasing the number of clusters. An exception to this rule arises when using the RN50x64 backbone on nuImages. Compared to other methods, using anchors shows lesser or on-par results, with the only exception being nuImages using RN50x4.
Using Attention for soft aggregation of embeddings (orange triangle) is beneficial for RN50 on all datasets and number of clusters; however, it greatly impairs performance for RN50x64.

\section{Cluster-OwlViT}

The Cluster-CLIP architecture is compatible with any dual-encoder VL open-vocabulary model, as elaborated in Section \ref{sec:related} of the main article. This section demonstrates this compatibility by implementing the Cluster-CLIP architecture with OwlViT backbones, referred to as Cluster-OwlViT. To achieve this, we apply the aggregation module outlined in Section \ref{sec:clustering} of the main article on top of OwlViT's dense embeddings.

\begin{figure}[b!]
{\includegraphics[width=\linewidth]{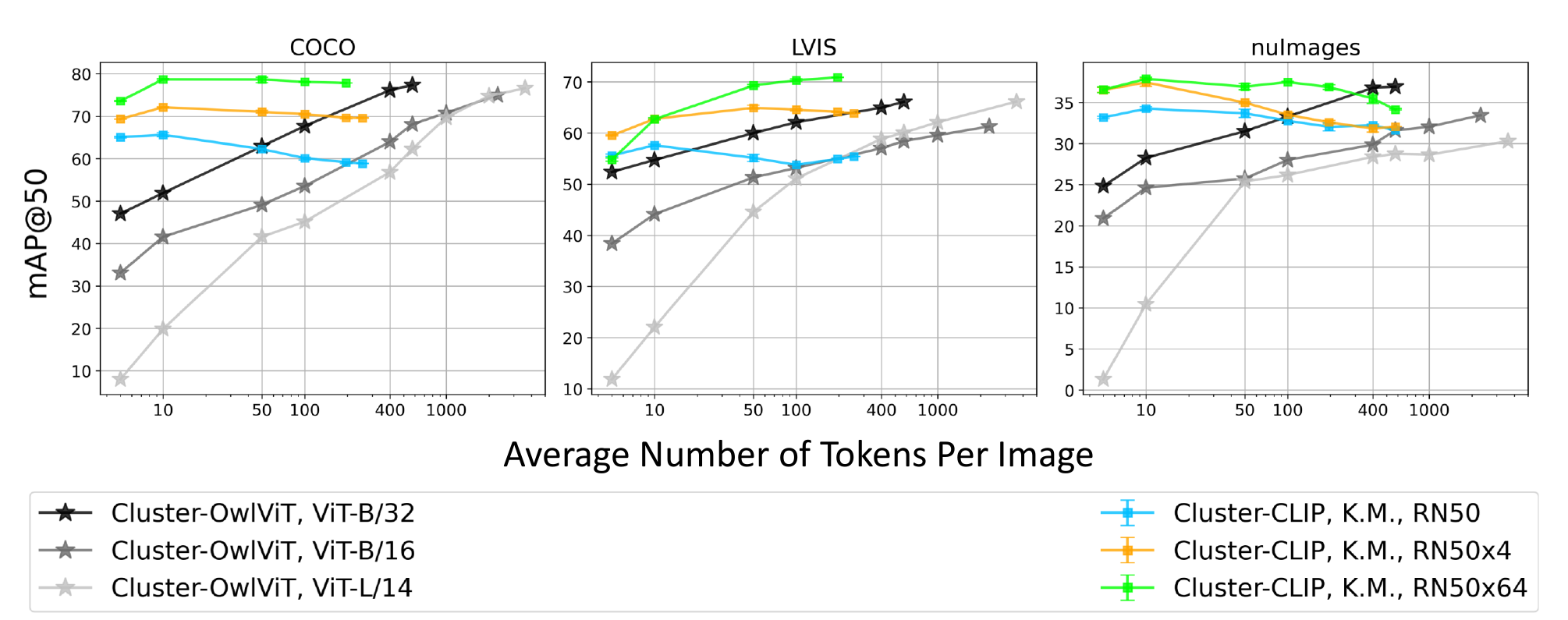}}\\
\caption{\textbf{Cluster-OwlViT accuracy-efficiency plots, compared to Cluster-CLIP-K.M.:} retrieval accuracy (mAP@50) vs. average numbers of embeddings per image for COCO, LVIS and nuImages datasets. Top left is better.} 
\vspace{-0.2cm}
\label{app:fig:cluster-owl-vit}
\end{figure}

Figure \ref{app:fig:cluster-owl-vit} presents the results of Cluster-OwlViT (represented by gray lines), compared to Cluster-CLIP using different backbones and K-Means clustering. When equipped with the ViT-B/32 backbone, Cluster-OwlViT maintains high retrieval rates while reducing the number of clusters by 30\% across all three datasets. With the larger ViT-L/14 backbone, Cluster-OwlViT remains competitive while managing to reduce the number of clusters from 3600 to 1000. As Cluster-CLIP outperforms Cluster-OwlViT with fewer representatives, we focus our work on it.

\section{Qualitative Examples}\label{app:qualitative_ex}

\begin{figure}[h]
{\includegraphics[width=\linewidth]{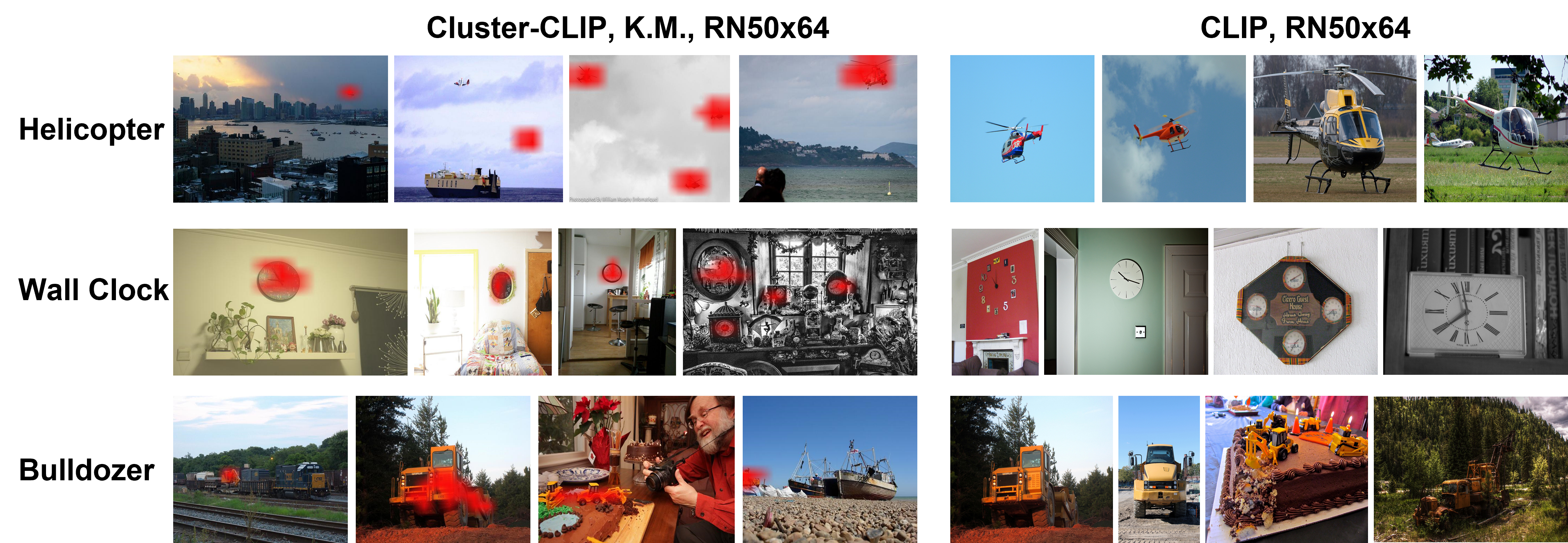}}\\
\caption{\textbf{Cluttered images, qualitative examples}. Using Cluster-CLIP representation (left part of the figure) allows the retrieval of cluttered images. High-scored patches are emphasized in red. Using global feature (right part) focuses on images with centered objects which occupy a large extent of the images. Best viewed in color while zoomed in.}
\label{fig:qual1}
\end{figure}

\begin{figure}[t!]
{\includegraphics[width=\linewidth]{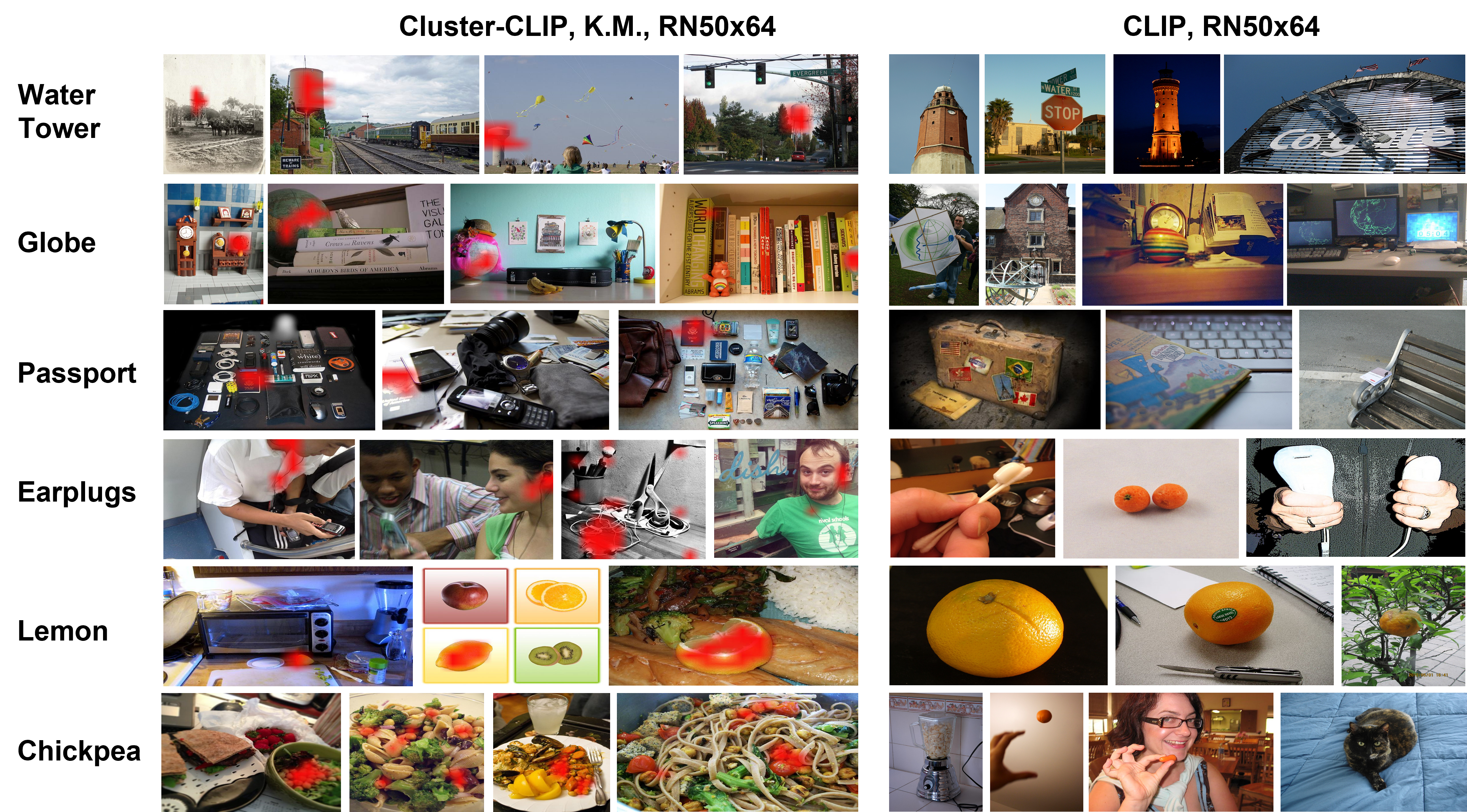}}\\
\caption{\textbf{Small objects, qualitative examples}. Using Cluster-CLIP representation (left side of the figure) allows the retrieval of images of relevant categories, even if the corresponding instances are relatively small. When global features are used (right side), images with centered and larger objects from semantically similar or even unrelated categories might score  higher. Best viewed in color while zoomed in.}
\label{fig:qual2}
\end{figure}

For demonstration purposes, and as discussed in Section \ref{sec:exp_qual} of the main article, we build two image retrieval framework indexes consisting of 120K COCO training set images. The first uses Cluster-CLIP-K.M. with RN50x64 backbone and 10 clusters, while the second uses CLIP with RN50x64 backbone. In both cases, FAISS \cite{johnson2019billion} is utilized to map the embeddings (aggregated embeddings for Cluster-CLIP, global embedding for CLIP) into a large-scale index. Qualitative retrieval examples of interest are presented in Figures \ref{fig:qual1} and \ref{fig:qual2}.

Figure \ref{fig:qual1} shows the top retrieval results for 'Helicopter', 'Wall clock', and 'Bulldozer' text queries. Using Cluster-CLIP allows the retrieval of cluttered images with relatively small instances of the requested category. Figure \ref{fig:qual2} shows top retrieval results for 'Water Tower', 'Globe', 'Passport', 'Earplugs', 'Lemon' and 'Chickpea' text queries, in which Cluster-CLIP produces desired results whereas CLIP prefers larger instances from false categories (sometimes semantically similar).  Both figures emphasize the importance of using non-global features for the object-centric image-retrieval task.  

\vspace{-0.5cm}

\section{Hyperparameters}

\noindent\textbf{Dense models.} 
We used the CLIP backbones (RN50, RN50x4, and RN50x64) from the CLIP  \cite{DBLP:conf/icml/RadfordKHRGASAM21} library and OwlViT framework \cite{owlvit} from the huggingface transformers library \cite{wolf-etal-2020-transformers} with default hyperparameters. Images were resized to a square aspect ratio (for details of the different resolutions, refer to Tables \ref{tab:dense_cocolvis} and \ref{tab:dense_nuimages} in the main article), and positional embeddings were interpolated to match the image resolution. For a fair comparison, we ensemble over the embeddings space of the 7 best CLIP prompts \cite{owlvit} in all baselines and experiments that use CLIP or OwlViT text encoders. 

\vspace{0.2cm}
\noindent\textbf{Clustering methods.}
We provide a detailed list of the different hyperparameters used in each of the clustering methods.
\begin{itemize}
    \item[--] K-Means - We used sklearn library \cite{scikit-learn} with the following configurations to run the K-Means clustering: \textit{init}=random, \textit{n\_init}=10, \textit{max\_iter}=300, \textit{tol}=0.0001, \textit{algorithm}=lloyd.
    
    \item[--] Region Proposals - We used Segment Anything library \cite{kirillov2023segany}, using the \textit{vit\_h} architecture along with its pre-trained weights, with different number of point-prompts (64, 256, 1024), an IoU threshold of 0.88, stability score threshold of 0.88, stability score offset of 0.1, box NMS threshold of 0.7, and no minimum mask region area nor running separately on crops of the image.  
    
    \item[--] Agglomerative Clustering - We used sklearn library with the following configurations to run Agglomerative clustering: \textit{linkage}=Ward, \textit{affinity}=Euclidean. Additionally, for Cluster-CLIP, AG-T, we set \textit{connectivity} to be a grid.
    
    \item[--] Adaptive K-Means -  The method attempts to select best number of clusters out of 5, 10, 15, and 20 clusters.
    
    \item[--] Anchors - We evaluated the following different divisions of the embeddings space: 
    \begin{inparaenum}[(1)]
        \item $1\times1$, $2\times2$
        \item $2\times2$, $3\times3$
        \item $3\times3$, $4\times4$, $5\times5$
        \item $2\times2$, $3\times3$, $5\times5$, $7\times7$
        \item $2\times2$, $3\times3$, $4\times4$, $5\times5$, $7\times7$
    \end{inparaenum}
    resulting in 5, 13, 50, 87 and 103 clusters respectively.
\end{itemize}

\pagebreak

\bibliography{arxiv/ar_eov}

\end{document}